\documentclass[11pt,a4paper]{article}

\usepackage[utf8]{inputenc}
\usepackage[T1]{fontenc}
\usepackage{lmodern}
\usepackage{amsmath,amssymb}
\usepackage{graphicx}
\usepackage[dvipsnames]{xcolor}
\usepackage{booktabs}
\usepackage{multirow}
\usepackage{array}
\usepackage{enumitem}
\usepackage{caption}
\usepackage{subcaption}
\usepackage{listings}
\usepackage[breaklinks,colorlinks,
            linkcolor=MidnightBlue,
            citecolor=MidnightBlue,
            urlcolor=MidnightBlue]{hyperref}
\usepackage{natbib}
\usepackage{geometry}
\usepackage{microtype}
\usepackage{float}

\title{\textbf{Can AI Draw Science? A Benchmark for Evaluating\\
Scientific Figure Generation by Text-to-Image\\ and Multimodal Models}}

\author{
  Davie Chen\thanks{E-mail: \href{mailto:davie@daviechen.com}{davie@daviechen.com}; ORCID: \href{https://orcid.org/0009-0001-4819-2828}{0009-0001-4819-2828}.}
}

\date{June 2026}

\begin{document}
\maketitle

\begin{abstract}
Text-to-image and multimodal generative models are increasingly used to produce scientific figures---mechanism diagrams, experimental-design schematics, conceptual frameworks, and graphical abstracts. Yet existing image-generation benchmarks (e.g., GenEval, T2I-CompBench, DPG-Bench) evaluate \emph{natural} images and measure compositionality, object counting, or photorealism. None of them measure the properties that determine whether a generated \emph{scientific} figure is usable: correct and legible text labels, faithful depiction of entities and their relations, coherent diagrammatic structure, and adherence to disciplinary drawing conventions. We introduce \textbf{SciDraw-Bench}, a benchmark of 32 structured scientific-figure generation tasks spanning eight figure types and ten disciplines, where each task pairs a natural-language prompt with a machine-checkable specification of required labels, relations, components, conventions, and negative constraints. We propose a four-dimensional evaluation protocol---Text Fidelity (OCR-based label recall and character error rate), Semantic Correctness (vision--language-model judging against the specification), Structural Quality, and Convention Adherence---together with a meta-evaluation protocol and a preliminary inter-judge reliability analysis (human-rating validation is ongoing). We evaluate a domain-specific system, \mbox{SciDraw AI}, against representative general-purpose text-to-image models, and outline a code-to-figure baseline (LLM-generated vector graphics) as a planned extension. The task specifications and evaluation harness are available from the author on request, supporting reproducible measurement of progress on scientific figure generation. In a pilot over all eight figure types, the domain-specific system substantially outperforms the general-purpose text-to-image baselines on every dimension and figure type, with the largest gaps on semantic correctness and disciplinary-convention adherence; text fidelity remains the hardest dimension for all systems.
\end{abstract}


\section{Introduction}
\label{sec:intro}

Scientific figures are a primary medium of scholarly communication: mechanism diagrams, experimental-design schematics, conceptual frameworks, and graphical abstracts compress complex ideas into a form that supports peer review, replication, and teaching~\citep{tufte2001visual,rougier2014ten}. Producing them well has traditionally required substantial time and specialized tooling. The rapid improvement of text-to-image (T2I) diffusion models~\citep{rombach2022high,saharia2022photorealistic}, autoregressive image models~\citep{yu2022scaling}, and multimodal foundation models~\citep{betker2023improving,team2023gemini} has made it plausible that such figures could be generated directly from natural-language descriptions, and a number of systems now target this use case.

Whether current models can actually draw \emph{science}, however, is an open empirical question---and one that existing benchmarks do not answer. The dominant T2I evaluations measure properties of \emph{natural} images: GenEval~\citep{ghosh2023geneval} and T2I-CompBench~\citep{huang2023t2i} probe object presence, counting, color binding, and spatial composition; DPG-Bench~\citep{hu2024ella} measures dense prompt following; HEIM~\citep{lee2023holistic} aggregates aesthetics, alignment, robustness, and bias; and FID~\citep{heusel2017gans} and CLIPScore~\citep{hessel2021clipscore} score distributional realism and image--text alignment. None of these capture what makes a scientific figure \emph{usable}:

\begin{enumerate}[label=(\roman*),nosep]
  \item \textbf{Text fidelity.} Scientific figures are label-dense. A mechanism diagram with garbled or misspelled gene names is worthless, yet T2I models are notoriously unreliable at in-image text~\citep{chen2023textdiffuser,yang2023glyphcontrol}.
  \item \textbf{Semantic correctness.} The right entities must appear in the right relations. ``PTEN inhibits PIP3'' must be drawn as inhibition, not activation.
  \item \textbf{Structural quality.} Reading order, arrow directionality, and grouping must communicate the intended logic.
  \item \textbf{Convention adherence.} Disciplines have visual grammars (membranes as bilayers, inhibition as blunt arrows, counter-current condenser flow) that domain readers expect.
\end{enumerate}

We address this gap with \textbf{SciDraw-Bench}. Our contributions are:

\begin{enumerate}[label=(\arabic*)]
  \item \textbf{A benchmark dataset} of 32 scientific-figure generation tasks across eight figure types and ten disciplines. Crucially, the dataset consists of \emph{tasks}, not reference images: each task is a prompt paired with a machine-checkable specification (required labels, relations, components, conventions, negative constraints). There is no single ground-truth image, because a specification admits many correct renderings (Section~\ref{sec:bench}).
  \item \textbf{An evaluation protocol} tailored to scientific figures, combining OCR-based Text Fidelity, vision--language-model (VLM) judging of Semantic Correctness, and rubric-based Structural Quality and Convention Adherence (Section~\ref{sec:protocol}).
  \item \textbf{A meta-evaluation protocol} for validating the automatic metrics against human expert ratings using rank correlation and inter-annotator agreement---the step that distinguishes a measurement instrument from an ad-hoc tool comparison---together with a preliminary inter-judge reliability analysis (Section~\ref{sec:protocol}); the human-rating study is ongoing.
  \item \textbf{A pilot comparative study} of a domain-specific system against representative general-purpose text-to-image baselines, with a qualitative error taxonomy (Sections~\ref{sec:setup}--\ref{sec:error}).
\end{enumerate}

The task specifications and evaluation harness are available from the author on request so that the benchmark can be re-run as models evolve. This manuscript fixes the benchmark and protocol and reports a pilot evaluation across all eight figure types (Sections~\ref{sec:results}--\ref{sec:error}); the human meta-evaluation is ongoing.

\section{Related Work}
\label{sec:related}

\paragraph{Text-to-image evaluation.}
Early evaluation of generative image models relied on distributional metrics such as FID~\citep{heusel2017gans} and on image--text alignment scores such as CLIPScore~\citep{hessel2021clipscore}. As prompt following became the bottleneck, prompt-centric benchmarks emerged: PartiPrompts~\citep{yu2022scaling}, GenEval~\citep{ghosh2023geneval}, T2I-CompBench~\citep{huang2023t2i}, and DPG-Bench~\citep{hu2024ella} measure object presence, counting, attribute binding, spatial relations, and dense prompt adherence. HEIM~\citep{lee2023holistic} broadens evaluation to multiple human-aligned dimensions. All of these target natural-image generation; none define a notion of \emph{diagrammatic correctness} or measure in-image scientific text, which is the focus of SciDraw-Bench.

\paragraph{Text rendering in images.}
The difficulty of rendering legible text has motivated specialized methods such as TextDiffuser~\citep{chen2023textdiffuser} and GlyphControl~\citep{yang2023glyphcontrol}, and corresponding evaluations based on OCR accuracy. We adopt an OCR-based label-recall metric in the same spirit, but apply it to the domain-specific labels demanded by a figure specification rather than to free-form text rendering.

\paragraph{Model- and VLM-as-judge.}
Using strong models to judge generated outputs is now standard for open-ended tasks~\citep{zheng2023judging}. We use a VLM-as-judge ensemble for the semantic and structural dimensions, force structured per-item verdicts against an explicit rubric, and report inter-judge agreement as a reliability check; validation against human ratings is part of our protocol and is ongoing.

\paragraph{Scientific visualization and tools.}
Principles of effective scientific figures are long established~\citep{tufte2001visual,rougier2014ten}, and tools such as BioRender and domain-specific AI platforms~\citep{scidraw2025} aim to lower the cost of producing them. Prior policy-oriented work has surveyed the editorial acceptability of AI-generated figures~\citep{chen2026aifigures}; here we instead ask the orthogonal, measurable question of how \emph{good} current systems are at the task.

\section{SciDraw-Bench}
\label{sec:bench}

\subsection{Design principle: tasks, not images}

A scientific figure specification is satisfied by many distinct images. A correct CAR-T mechanism diagram may place the T cell on the left or the right, use different iconography for the receptor, and still be fully correct. Pinning a single reference image as ground truth would therefore penalize valid variation and reward stylistic mimicry. SciDraw-Bench instead defines each item as a \emph{generation task}: a natural-language prompt (the input given verbatim to every system) plus a structured \emph{specification} of what any correct output must contain. Evaluation scores a generated image against the specification, not against a reference image.

\subsection{Task schema}

Each task is a JSON record with the following fields:

\begin{lstlisting}
{
  "id": "mech-001",
  "figure_type": "mechanism",
  "discipline": "immunology",
  "difficulty": "easy",
  "prompt": "Draw a molecular mechanism diagram of a CAR-T cell ...",
  "required_labels": ["CAR-T cell","HER2","scFv","CD3z", ...],
  "required_relations": [["scFv","binds","HER2"], ...],
  "required_components": ["T-cell with membrane bilayer", ...],
  "discipline_conventions": ["receptor as Y-shaped transmembrane", ...],
  "negative_constraints": ["no photorealistic micrographs", ...]
}
\end{lstlisting}

Each field maps to an evaluation dimension (Section~\ref{sec:protocol}): \texttt{required\_labels}~$\rightarrow$ Text Fidelity; \texttt{required\_relations} and \texttt{required\_components}~$\rightarrow$ Semantic Correctness; layout/ordering cues in the prompt~$\rightarrow$ Structural Quality; \texttt{discipline\_conventions} and \texttt{negative\_constraints}~$\rightarrow$ Convention Adherence.

\subsection{Coverage and statistics}

SciDraw-Bench contains 32 tasks, balanced as four tasks for each of eight figure types (Table~\ref{tab:coverage}). Each figure type contains one \emph{easy}, two \emph{medium}, and one \emph{hard} task, where difficulty scales with the number of required labels, the density of required relations, and the strictness of conventions (8 easy / 16 medium / 8 hard overall). Tasks span ten disciplines---biomedicine, cell biology, immunology, chemistry, physics, computer science, environmental science, materials science, neuroscience, and social science---so that no single visual idiom dominates. The set is intentionally compact and fully hand-authored to keep specifications accurate and auditable; the schema and harness are designed so the suite can be extended without changing the protocol.

\begin{table}[htbp]
\centering
\caption{SciDraw-Bench figure types (4 tasks each; 32 total). Each type spans easy/medium/hard difficulty.}
\label{tab:coverage}
\small
\begin{tabular}{@{}llp{6.4cm}@{}}
\toprule
\textbf{Figure type} & \textbf{ID prefix} & \textbf{What it stresses} \\
\midrule
Mechanism / pathway       & \texttt{mech} & dense labels, activation/inhibition relations \\
Experimental design       & \texttt{exp}  & groups, timeline, endpoints, structured layout \\
Conceptual framework      & \texttt{frame}& hierarchy and grouping of objectives/methods \\
Technical roadmap         & \texttt{road} & milestones, temporal ordering \\
Graphical abstract        & \texttt{gabs} & one-glance composition and balance \\
Pipeline / flowchart      & \texttt{pipe} & step ordering, arrow directionality \\
Apparatus schematic       & \texttt{appa} & spatial fidelity of physical setups \\
Concept map               & \texttt{cmap} & entities and labeled relations \\
\bottomrule
\end{tabular}
\end{table}

\section{Evaluation Protocol}
\label{sec:protocol}

We score each generated image on four dimensions, each normalized to $[0,1]$, plus a composite. For each (system, task) pair we generate $k{=}3$ samples to account for stochasticity and report both best-of-$k$ and mean. All model versions and generation timestamps are logged for reproducibility.

\subsection{Text Fidelity (TF)}
We extract the text tokens present in the image by OCR; the harness supports PaddleOCR~\citep{du2020ppocr} and Tesseract backends, and the pilot reported here uses a vision--language-model transcription pass. For each string in \texttt{required\_labels}, we test for a fuzzy match against the OCR tokens using a normalized Levenshtein distance (we accept a match when the normalized distance is below a threshold $\tau$; the harness uses $\tau{=}0.30$). We report \emph{label recall} (fraction of required labels found) and the mean character error rate over matched labels, and combine them into
\begin{equation}
\mathrm{TF} = \alpha \cdot \text{recall} + (1-\alpha)\,(1-\text{CER}), \qquad \alpha = 0.7 .
\end{equation}

\subsection{Semantic Correctness (SC)}
A VLM-as-judge is shown the generated image and the task specification and returns a per-item yes/no verdict for each entry of \texttt{required\_relations} and \texttt{required\_components}: is the relation depicted with the correct directionality, is the component present, and are there hallucinated entities that contradict the specification? SC is the fraction of specification items satisfied. To reduce single-judge bias we use an ensemble of at least two judge models and average their item-level verdicts. The exact rubric prompt is fixed and included in the harness so that judgments are auditable and reportable.

\subsection{Structural Quality (SQ) and Convention Adherence (CA)}
SQ is a VLM-judged 1--5 rubric covering reading order, arrow directionality, grouping coherence, and absence of overlap clutter. CA is a VLM-judged 1--5 rubric checking the task's \texttt{discipline\_conventions} and penalizing violations of \texttt{negative\_constraints}. Both are linearly mapped to $[0,1]$.

\subsection{Composite}
The composite score is a weighted mean
\begin{equation}
\mathrm{Overall} = w_{\mathrm{TF}}\mathrm{TF} + w_{\mathrm{SC}}\mathrm{SC} + w_{\mathrm{SQ}}\mathrm{SQ} + w_{\mathrm{CA}}\mathrm{CA},
\end{equation}
with default weights $(0.30,0.30,0.20,0.20)$. We report a sensitivity analysis over weights in Appendix~\ref{app:weights}.

\subsection{Meta-evaluation}
Automatic metrics are only useful if they track human judgment. The protocol therefore collects human ratings from at least three annotators on a stratified subset of generated images, on the same four dimensions (1--5), and reports (i) Spearman $\rho$ and Kendall $\tau$ between automatic and mean-human scores per dimension, and (ii) inter-annotator agreement via Krippendorff's $\alpha$ and ICC(2,$k$). Strong rank correlation and acceptable agreement are required before the automatic scores are interpreted as calibrated measurements rather than estimates. This human study is ongoing; as an interim reliability check we report inter-judge agreement between the two automatic judges (Section~\ref{sec:meta}).

\section{Experimental Setup}
\label{sec:setup}

\paragraph{Systems under test.}
We compare a domain-specific scientific-illustration platform, \mbox{SciDraw AI} (\url{https://sci-draw.com})~\citep{scidraw2025}, against representative general-purpose text-to-image models (Table~\ref{tab:systems}). Because SciDraw AI is a model-agnostic platform---its visual backend is configurable---we compare it against standalone text-to-image models rather than treating any single foundation model as a fixed rival. The question is whether the platform's domain tooling helps on the dimensions general models struggle with. A code-to-figure baseline (an LLM emitting SVG/TikZ/Mermaid, subsequently rendered) and additional diffusion models are planned extensions of the suite.

\begin{table}[htbp]
\centering
\caption{Systems evaluated in the pilot. Specific model versions and generation dates are logged in the run configuration.}
\label{tab:systems}
\small
\begin{tabular}{@{}lp{7.6cm}@{}}
\toprule
\textbf{System} & \textbf{Class} \\
\midrule
SciDraw AI    & Domain-specific scientific-illustration platform \\
Ideogram v3   & Text-specialized text-to-image model \\
Qwen-Image    & Open general-purpose text-to-image model \\
\bottomrule
\end{tabular}
\end{table}

\paragraph{Protocol controls.}
Every system receives the identical task prompt; we apply no benchmark-side prompt engineering. The domain-specific platform additionally applies its own production scientific-illustration instruction layer, which is a property of the system under test rather than a benchmark intervention. The pilot uses $k{=}2$ samples per task (the full protocol specifies $k{=}3$). The two judge models (GPT-4o and Gemini) are drawn from families disjoint from the text-to-image baselines; we additionally report inter-judge agreement (Section~\ref{sec:meta}) as a reliability check.

\section{Results}
\label{sec:results}

\begin{figure}[htbp]
  \centering
  \begin{subfigure}{0.48\textwidth}
    \includegraphics[width=\linewidth]{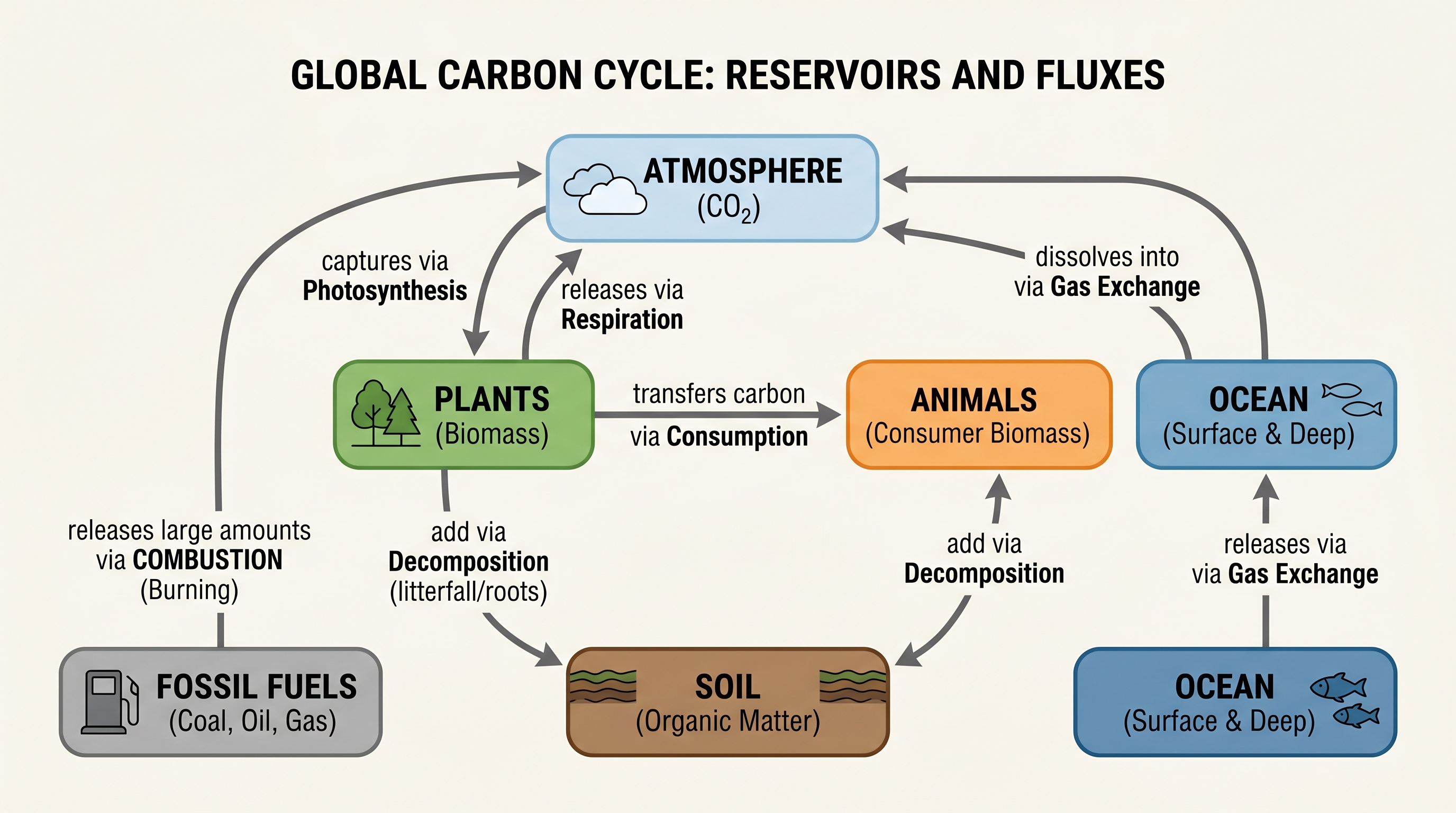}
    \caption{Concept map (\texttt{cmap-001}, carbon cycle)}
  \end{subfigure}\hfill
  \begin{subfigure}{0.48\textwidth}
    \includegraphics[width=\linewidth]{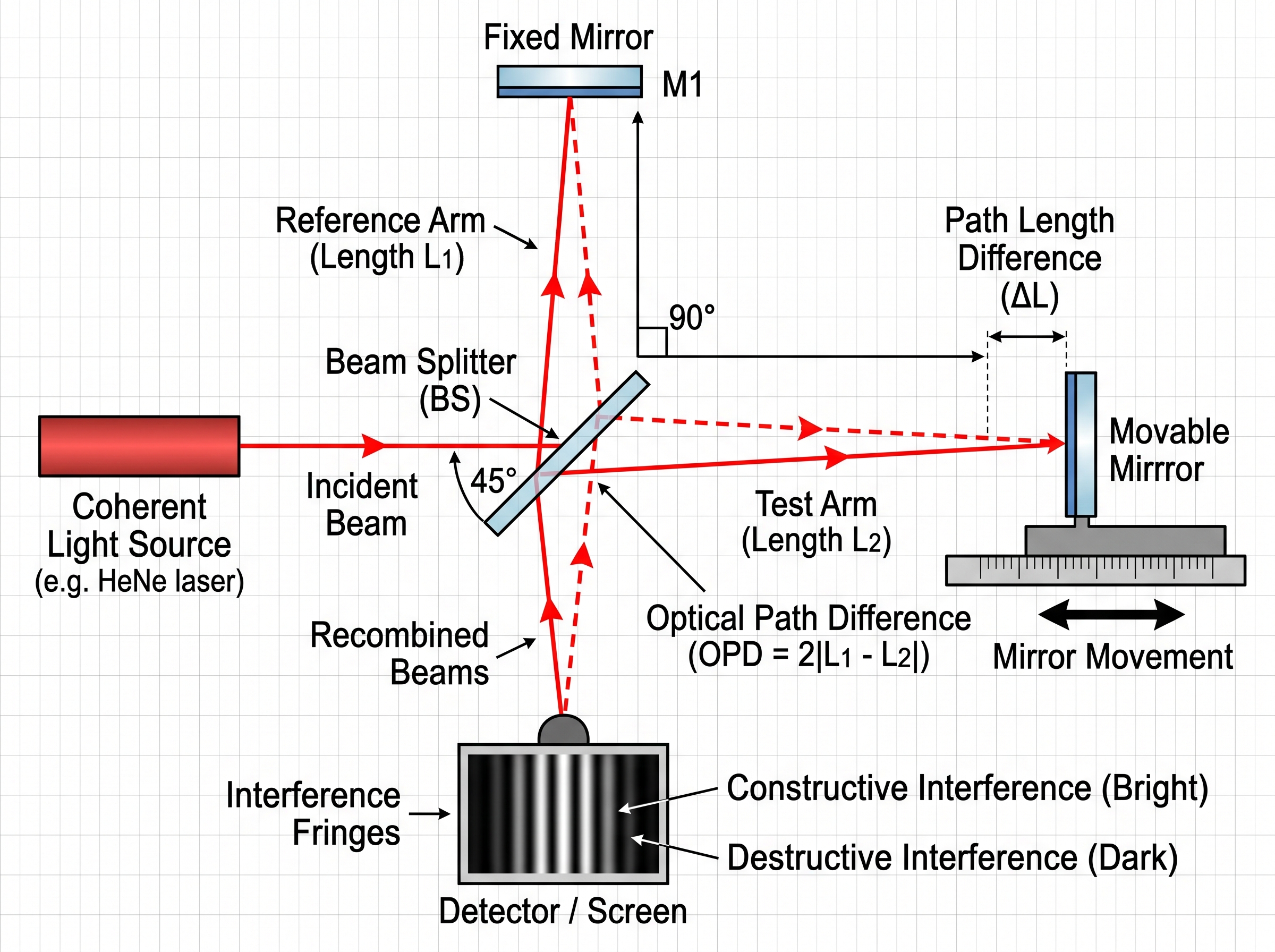}
    \caption{Apparatus schematic (\texttt{appa-002}, Michelson interferometer)}
  \end{subfigure}

  \medskip
  \begin{subfigure}{0.48\textwidth}
    \includegraphics[width=\linewidth]{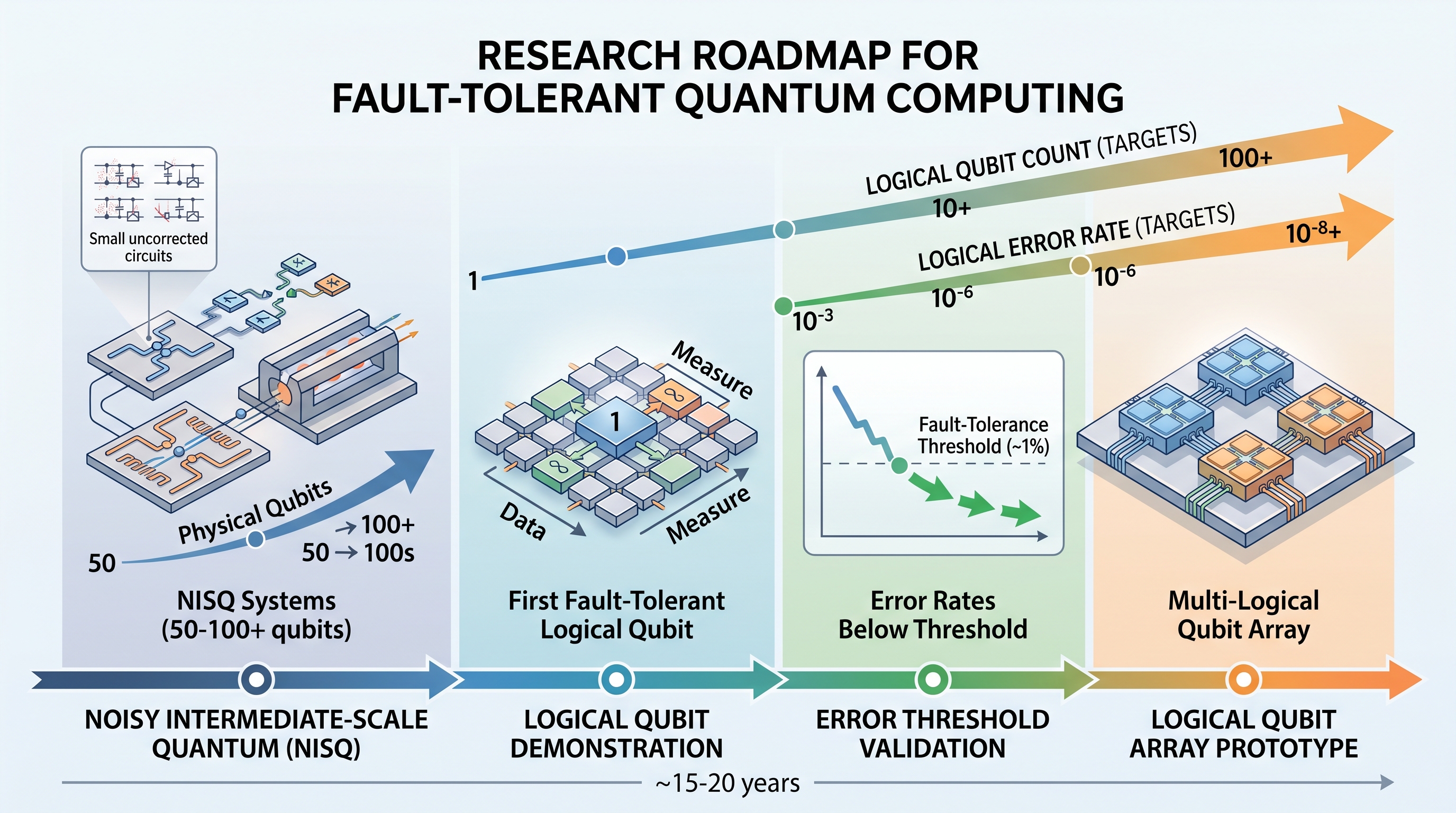}
    \caption{Technical roadmap (\texttt{road-003}, fault-tolerant QC)}
  \end{subfigure}\hfill
  \begin{subfigure}{0.48\textwidth}
    \includegraphics[width=\linewidth]{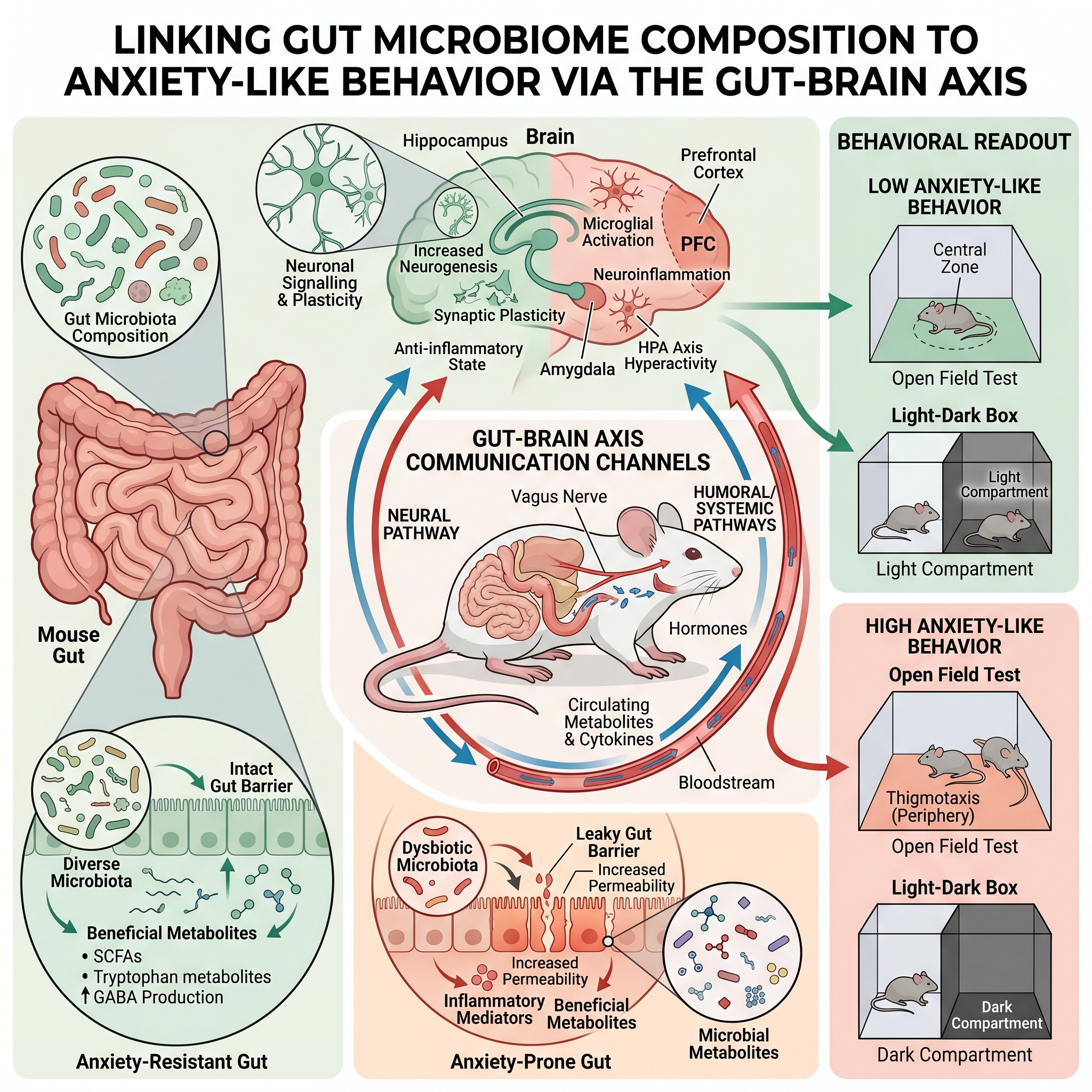}
    \caption{Graphical abstract (\texttt{gabs-002}, gut--brain axis)}
  \end{subfigure}
  \caption{Representative SciDraw-Bench tasks across four figure types, rendered by the domain-specific system (\mbox{SciDraw AI}). Labels, relations, and disciplinary conventions are largely satisfied. These are single ($k{=}1$) unedited outputs.}
  \label{fig:showcase}
\end{figure}

\paragraph{Qualitative pilot examples.}
Figure~\ref{fig:showcase} shows SciDraw-Bench tasks from four different figure types rendered by the domain-specific system, illustrating the range of scientific illustration the benchmark targets---concept maps, apparatus schematics, technical roadmaps, and graphical abstracts---and the level of text and convention fidelity attainable on these tasks.

\begin{table}[htbp]
\centering
\caption{Main results: mean dimension scores and composite score per system, averaged over $k{=}2$ samples on all 8 pilot tasks (16 images/system), scored by the two-judge ensemble. Best per column in bold.}
\label{tab:results}
\small
\begin{tabular}{@{}lccccc@{}}
\toprule
\textbf{System} & \textbf{TF} & \textbf{SC} & \textbf{SQ} & \textbf{CA} & \textbf{Overall} \\
\midrule
SciDraw AI    & \textbf{0.74} & \textbf{0.98} & \textbf{0.98} & \textbf{0.98} & \textbf{0.91} \\
Qwen-Image    & 0.61 & 0.80 & 0.59 & 0.51 & 0.64 \\
Ideogram v3   & 0.59 & 0.64 & 0.55 & 0.48 & 0.57 \\
\bottomrule
\end{tabular}
\end{table}

Table~\ref{tab:results} reports mean scores over the 16 images per system; Table~\ref{tab:bytype} (Appendix) breaks the composite down by figure type. Three findings stand out.

\textbf{The domain-specific platform dominates the general-purpose baselines.} SciDraw AI scores 0.91 on the composite, far above Qwen-Image (0.64) and Ideogram v3 (0.57). The separation is largest on the judged dimensions: the general-purpose models frequently omit or misorder required entities (SC $0.80$/$0.64$ vs $0.98$) and violate disciplinary conventions (CA $0.51$/$0.48$ vs $0.98$), and at times return decorative, off-task output with no diagrammatic structure, whereas SciDraw AI satisfies nearly all specification items.

\textbf{Text Fidelity is the universal bottleneck.} No system exceeds 0.74 on TF---markedly lower than the SC/SQ/CA scores---confirming that rendering correct, legible in-image scientific text remains the hardest part of the task even for the strongest system (cf.\ the garbled and misspelled labels in Figure~\ref{fig:failure}).

\textbf{The advantage holds across every figure type.} SciDraw AI achieves the highest composite on all eight figure types (Table~\ref{tab:bytype}), with its largest margins on communication-oriented figures---concept maps ($1.00$), conceptual frameworks ($0.95$), graphical abstracts ($0.97$), and technical roadmaps ($0.89$). This matches our pre-registered expectation (H3) that domain-specific tooling helps most on template-like, communicative figures, and that general T2I models score lowest on Text Fidelity (H1).

\section{Meta-Evaluation}
\label{sec:meta}

A full meta-evaluation validates the automatic metrics against human expert ratings; collecting those ratings is ongoing and we report it as future work. As a reliability check available now, Table~\ref{tab:meta} reports agreement between the two independent judges (OpenAI GPT-4o and Gemini) on the three judged dimensions over the $n{=}46$ scored image pairs. Agreement is moderate (Pearson $r$ between 0.58 and 0.63; mean absolute difference $\le 0.21$ on a $[0,1]$ scale), which supports averaging them as an ensemble while motivating the human validation below. Human-validated correlations remain to be established before the automatic scores are interpreted as calibrated measurements rather than internally consistent estimates.

\begin{table}[htbp]
\centering
\caption{Inter-judge agreement between the two independent judges on the judged dimensions ($n{=}46$ image pairs). Text Fidelity is computed programmatically from OCR and is not judge-scored. Human meta-evaluation is future work.}
\label{tab:meta}
\small
\begin{tabular}{@{}lccc@{}}
\toprule
\textbf{Dimension} & \textbf{Pearson $r$} & \textbf{Exact agree.} & \textbf{Mean abs.\ diff.} \\
\midrule
Semantic Correctness & 0.62 & 0.43 & 0.14 \\
Structural Quality   & 0.58 & 0.52 & 0.19 \\
Convention Adherence & 0.63 & 0.54 & 0.21 \\
\bottomrule
\end{tabular}
\end{table}

\section{Error Analysis}
\label{sec:error}

\begin{figure}[htbp]
  \centering
  \begin{subfigure}{0.48\textwidth}
    \includegraphics[width=\linewidth]{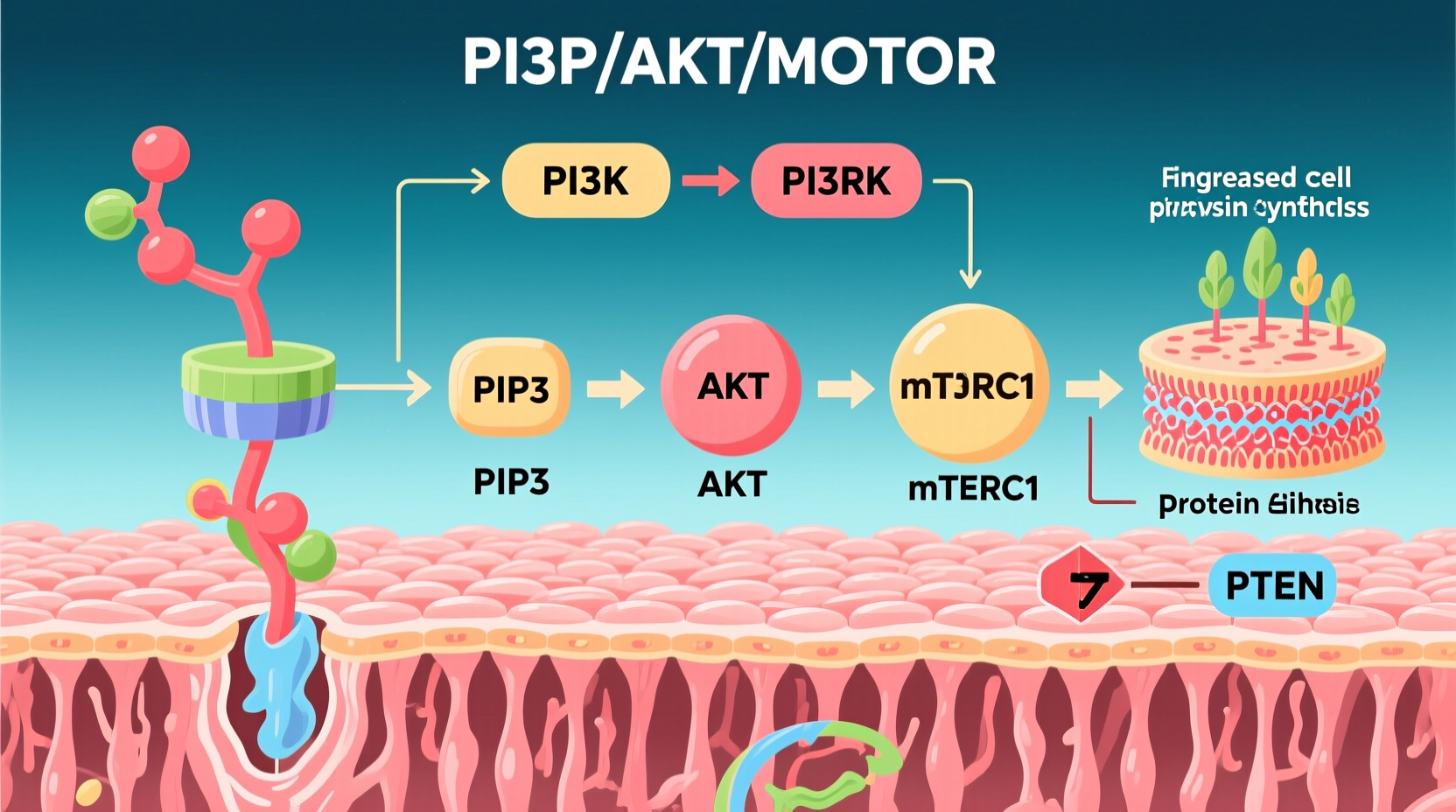}
    \caption{Garbled text (Qwen-Image)}
  \end{subfigure}\hfill
  \begin{subfigure}{0.48\textwidth}
    \includegraphics[width=\linewidth]{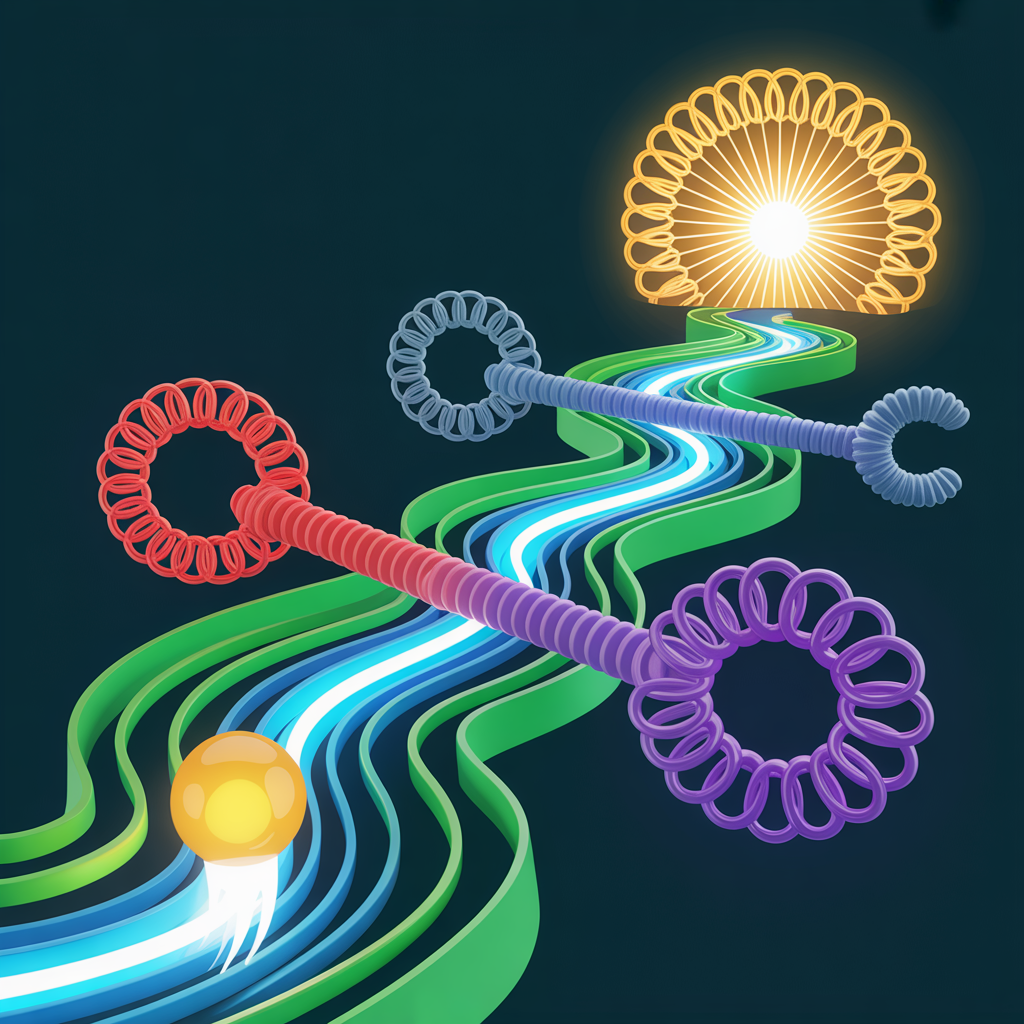}
    \caption{Off-task rendering (Ideogram v3)}
  \end{subfigure}
  \caption{Characteristic failure modes of general-purpose text-to-image models on the identical task \texttt{mech-002} (PI3K/AKT/mTOR signaling), single ($k{=}1$) unedited outputs: (a) a diagram-like layout with severely garbled, misspelled labels, and (b) an off-task decorative abstraction lacking any diagrammatic structure or text.}
  \label{fig:failure}
\end{figure}

The general-purpose baselines fail in characteristic ways, illustrated in Figure~\ref{fig:failure} on the task \texttt{mech-002} (the PI3K/AKT/mTOR cascade). One model produces a plausibly laid-out diagram whose text is badly garbled (e.g., ``PI3P/AKT/MOTOR'', ``mTJRC1'', ``mTERC1'') with misspelled and duplicated labels; the other returns an off-task decorative abstraction with no diagrammatic content or labels at all. By contrast, the domain-specific system renders the same pathway with all required labels (RTK, PI3K, PIP$_2$/PIP$_3$, PTEN, AKT, mTORC1, S6K1, 4E-BP1) legible and correctly spelled, appropriate activation/inhibition glyphs, and a membrane bilayer (the comparable mechanism output appears among the examples in Figure~\ref{fig:showcase} for other figure types).

These observations motivate the failure taxonomy the full evaluation quantifies: (i) \emph{garbled text}---misspelled or unreadable labels (Text Fidelity); (ii) \emph{off-task rendering}---decorative output lacking diagrammatic structure (Structural Quality); (iii) \emph{hallucinated entities}---elements absent from the specification (Semantic Correctness); (iv) \emph{relation inversion}---inhibition drawn as activation or reversed arrows (Semantic Correctness); and (v) \emph{convention violation}---e.g., membranes not drawn as bilayers (Convention Adherence). The automatic metrics in Section~\ref{sec:protocol} are constructed to score exactly these properties, and each flagged example is linked to the specific specification item it violates, making the failure mode reproducible rather than anecdotal. We caution that these are single-task illustrations; per-system rankings (Section~\ref{sec:results}) are established by the aggregate scores over all 16 images per system, and the human-validated meta-evaluation remains ongoing.

\section{Discussion}
\label{sec:discuss}

SciDraw-Bench reframes ``can AI draw science?'' as a measurable question with per-dimension answers rather than a holistic impression. By scoring against specifications rather than reference images, it credits valid visual variation while still enforcing correctness. The four dimensions also map onto distinct intervention points: Text Fidelity is bounded by in-image text rendering; Semantic Correctness by prompt grounding; and Convention Adherence by domain knowledge---suggesting that hybrid or domain-specialized pipelines may dominate on different axes. The results connect directly to editorial guidance on AI-generated figures~\citep{chen2026aifigures}: the dimension on which every system remains weakest---Text Fidelity---is precisely the one where human review and disclosure remain non-negotiable, since even the strongest models still misspell or garble in-image labels.

\section{Limitations}
\label{sec:limits}

Several limitations apply. (i) \emph{Pilot scale}: the suite contains 32 tasks; the evaluation reported here is a pilot over 8 tasks (one per figure type) and 3 systems at $k{=}2$, giving broad type coverage but limited statistical power within cells. The protocol is designed to scale to the full suite and more systems without change. (ii) \emph{Non-exhaustive comparison}: we compare against widely-available general-purpose text-to-image baselines; this is not an exhaustive comparison, and strong commercial multimodal image models can match domain-specific output quality on several figure types. A controlled comparison against such models, and against the planned code-to-figure baseline, is left to future work. (iii) \emph{Stochasticity}: we mitigate but do not eliminate sampling variance via $k{=}2$ and mean reporting. (iv) \emph{Judge bias and saturation}: VLM judges may exhibit systematic leniency; the near-perfect judged scores of the leading system suggest the 1--5 rubrics saturate at the high end and may under-resolve differences between strong systems. We use a two-judge ensemble and report inter-judge agreement, but human validation remains future work. (v) \emph{Specification subjectivity}: ``correct'' conventions encode authorial judgment; specifications are available for scrutiny and revision. (vi) \emph{Temporal validity}: model versions change quickly, so all results are timestamped and the harness is available for re-running.

\section{Conclusion}
\label{sec:conclusion}

We introduced SciDraw-Bench, the first benchmark, to our knowledge, dedicated to scientific figure generation, together with a four-dimensional evaluation protocol and a meta-evaluation protocol for validating the automatic metrics against human judgment (with a preliminary inter-judge reliability analysis). The benchmark measures the properties that decide whether a generated figure is usable in research---text fidelity, semantic correctness, structure, and convention---that natural-image benchmarks omit. The task specifications and evaluation harness are available from the author on request, and we invite the community to extend the suite and re-run it as models improve.

\section*{Data and code availability}
The SciDraw-Bench task specifications, the evaluation harness, the exact judge rubric prompts, and the generated artifacts and run configurations are available from the author on reasonable request.

\bibliographystyle{plainnat}

\appendix
\section{Weight sensitivity}
\label{app:weights}
We recompute the composite under alternative weightings to confirm that conclusions are not artifacts of the default weights $(0.30,0.30,0.20,0.20)$. SciDraw AI scores $0.92$ under balanced weights $(0.25,0.25,0.25,0.25)$ and $0.86$ under text-heavy weights $(0.5,0.3,0.1,0.1)$, versus $0.63$/$0.66$ for Qwen-Image and $0.57$/$0.59$ for Ideogram v3. The ranking (SciDraw AI $\gg$ Qwen-Image $>$ Ideogram v3) is preserved under all three weightings.

\section{Per-type breakdown}
\label{app:bytype}
\begin{table}[htbp]
\centering
\caption{Composite score by figure type (mean over $k{=}2$ samples, ensemble judges). Best per row in bold. SciDraw AI leads on every figure type, by the widest margins on communication-oriented figures.}
\label{tab:bytype}
\small
\begin{tabular}{@{}lccc@{}}
\toprule
\textbf{Figure type} & \textbf{SciDraw AI} & \textbf{Qwen-Image} & \textbf{Ideogram v3} \\
\midrule
Mechanism            & \textbf{0.92} & 0.55 & 0.28 \\
Experimental design  & \textbf{0.88} & 0.74 & 0.48 \\
Conceptual framework & \textbf{0.95} & 0.56 & 0.55 \\
Technical roadmap    & \textbf{0.89} & 0.62 & 0.81 \\
Graphical abstract   & \textbf{0.97} & 0.78 & 0.75 \\
Pipeline / flowchart & \textbf{0.81} & 0.59 & 0.61 \\
Apparatus schematic  & \textbf{0.87} & 0.74 & 0.42 \\
Concept map          & \textbf{1.00} & 0.59 & 0.69 \\
\bottomrule
\end{tabular}
\end{table}

\end{document}